\documentclass[runningheads]{llncs}
\usepackage{graphicx}
\usepackage{tabularx}
\usepackage{booktabs}
\usepackage{epstopdf}

\begin{document}
\title{PuoBERTa: Training and evaluation of a curated language model for Setswana}
%
\titlerunning{PuoBERTa: A curated language model for Setswana}
%
\author{Vukosi Marivate\inst{1,2}\orcidID{0000-0002-6731-6267} \and
Moseli Mots'Oehli\inst{3}\and
 Valencia Wagnerinst{4}\orcidID{0000-0003-2671-7512}\and
Richard Lastrucci \inst{1}\and
Isheanesu Dzingirai \inst{1}}
\authorrunning{V. Marivate et al.}
%
\institute{Department of Computer Science, University of Pretoria \email{vukosi.marivate@cs.up.ac.za}\and Lelapa AI \and  University of Hawaii at Manoa \\ \email{moselim@hawaii.edu} \and Sol Plaatje University \\ \email{valencia.wagner@spu.ac.za}}
\maketitle              
\begin{abstract}
Natural language processing (NLP) has made significant progress for well-resourced languages such as English but lagged behind for low-resource languages like Setswana. This paper addresses this gap by presenting PuoBERTa, a customised masked language model trained specifically for Setswana. We cover how we collected, curated, and prepared diverse monolingual texts to generate a high-quality corpus for PuoBERTa's training. Building upon previous efforts in creating monolingual resources for Setswana, we evaluated PuoBERTa across several NLP tasks, including part-of-speech (POS) tagging, named entity recognition (NER), and news categorisation. Additionally, we introduced a new Setswana news categorisation dataset and provided the initial benchmarks using PuoBERTa. Our work demonstrates the efficacy of PuoBERTa in fostering NLP capabilities for understudied languages like Setswana and paves the way for future research directions. 
\end{abstract}



\section{Introduction}

The development of monolingual models for local languages is essential in better understanding individual languages and further developing tools for those languages \cite{limisiewicz-etal-2023-data}. It has been shown that monolingual models can further be improved for downstream tasks later on \cite{de-vries-etal-2021-adapting,baziotis2023monolingual} or even allow for cross-lingual transfer \cite{souza2021ability} and also have a space in improving translation models \cite{burlot-yvon-2018-using}. Ultimately, the availability of more monolingual models and the curation of data to train them will increase the opportunities for bigger multilingual models \cite{doddapaneni2022towards}, as they will have a larger pool of data to learn from, better evaluation for the individual languages and knowledge of the limitations of approaches and knowledge of the various limitations of approaches utilised by monolingual models (e.g. the development of subword segmentation to improve language models for Nguni languages \cite{meyer2022subword}). Additionally, monolingual models will be better suited for tasks that require a deep understanding of a specific language \cite{armengol-estape-etal-2021-multilingual}.

In this work, we focus our efforts on creating a monolingual language model for Setswana. Setswana is a Bantu languages that is spoken in Botswana as well as several regions of South Africa \cite{palai2004word}. Previous work has been done on creating monolingual datasets for Setswana (see NCHLT corpus \cite{eiselen2014developing}). This work has been valuable in providing a foundation for further research on Setswana. More recently the NCHLT corpus has been used to train language models that are available online without benchmarks\footnote{\url{http://www.rma.nwu.ac.za/handle/20.500.12185/641}} (as per the writing of this paper). The creation of more models provides diversity in approaches and benchmarks, which can help us to better evaluate the quality of resources available for a given language and understand the strengths and weaknesses of different techniques to natural language processing for Setswana. This paper contributes to the Setswana literature and also provides tools for researchers, and identifies areas where more research is needed. 

PuoBERTa is a customised masked language model trained specifically for Setswana. It was developed to address the gap in natural language processing (NLP) performance for Setswana. This paper covers how we collected, curated, and prepared diverse multilingual texts to generate a high-quality corpus for PuoBERTa's training. We also built upon previous efforts in creating monolingual resources for Setswana, and evaluated PuoBERTa across several NLP tasks, including part-of-speech (POS) tagging, named entity recognition (NER), and news categorization. Additionally, we introduce a new Setswana news categorization dataset and provide the initial benchmarks using PuoBERTa.

\section{Related Work}

There have been a few attempts to create a monolingual Setswana language model. TswanaBert~\footnote{\url{https://huggingface.co/MoseliMotsoehli/TswanaBert}} \cite{TshwanaBert} is one of the earliest models available for Setswana.  It was trained on a combination of datasets from scraped news headlines from social media\footnote{\url{https://zenodo.org/record/5674236}} \cite{marivate-etal-2020-investigating} and the Leipzig Corpus \footnote{\url{https://corpora.uni-leipzig.de/}} \cite{goldhahn-etal-2012-building,tsn-za_web_2019}. 

A more recent model based on the NCHLT \cite{eiselen2014developing}, Autshumato, Leipzig and Webcrawl corpora is the NCHLT RoBERTa Model\footnote{\url{http://www.rma.nwu.ac.za/handle/20.500.12185/641}} \cite{nchlt-RoBERTa-tsn}. For both TswanaBert and NCHLT Setswana RoBERTa language models, at the time of writing, there are no online benchmarks that compare the approaches to other work, such as multilingual models that include the Setswana language. As such we later provide benchmarks for these models on Setswana downstream tasks. 

More recently, work on African language models has focused on creating multilingual models. The multilingual approaches aim to exploit the connections between African languages to allow languages with low resources to benefit from those that have more resources. Work by Adelani et.al.\cite{adelani-etal-2022-thousand} focused on creating parallel corpora, for machine translation, for a number of African languages including Setswana. The corpora is then used to fine-tune pre-trained translation models such as MT5 \cite{xue2020mt5}, MBART \cite{liu2020multilingual}, ByT5 \cite{xue2021byt5} and M2M-100 \cite{fan2020englishcentric}. Other work focused on adapting XLM-RoBERTa into African languages including Setswana \cite{alabi-etal-2022-adapting}, training multilingual BERT for African languages \cite{ogueji-etal-2021-small}. Additionally, with the advent of the age of massive large language models, we have models such as BLOOM \cite{scao2022language} which have Setswana within and more Afrocentric models such as Serengeti \cite{adebara-etal-2023-serengeti} that are now available.  

\section{Curating Setswana Corpora {PuoData} and Pre-Training PuoBERTa}

\subsection{Curating Corpora to Train Word and Sentence Representations}

Modern language models require a large amount of data to develop. The challenges of obtaining data for low-resource languages have been well-documented \cite{ragni2014data,ranathunga2021neural,nekoto-etal-2020-participatory}. In this work, we gather a number of datasets to pre-train Setswana language models. The data was includes data from research organisations, books, official government documents, and online content. The final dataset is referred to as PuoData. Puo means "language" in Setswana.

We believe that PuoData is a valuable resource for the Setswana language community. We hope that PuoData will be used to develop new and innovative applications that benefit the Setswana-speaking community. One of the input corpora that we gather during this project (starting in 2020) was the JW300 monolingual dataset for Setswana from the OPUS Corpora \cite{aulamo-etal-2020-opustools}. However, JW300 is no longer an open source dataset available online. As such, we will refer to PuoData+JW300, which is the dataset including a historical JW300 archive we had access to before it was removed from the OPUS. We provide benchmarks with  PuoData and PuoData+JW300 in the interest of language development. PuoData+JW300 is a larger dataset than PuoData alone (see Table \ref{tab:datasets}). It contains more text (even if it is text that is religious in nature) than all the other datasets combined. 

The data, detailed in Table \ref{tab:datasets}, contains sources such as the NCHLT Setswana \cite{eiselen2014developing} corpus, the South African Constitution \footnote{\url{https://www.justice.gov.za/constitution/SAConstitution-web-set.pdf}}, the Leipzig Setswana BW and ZA corpora. We also include more recent corpora such as the Vuk'zenzele Setswana Corpora \cite{lastrucci2023preparing}\footnote{\url{https://huggingface.co/datasets/dsfsi/vukuzenzele-monolingual}} and South African Cabinet Speeches \cite{lastrucci2023preparing}\footnote{\url{https://huggingface.co/datasets/dsfsi/gov-za-monolingual}}. All of the corpora contain Setswana data only

\begin{table}
\centering
\label{tab:datasets}
\caption{Setswana monolingual curated corpora with relative sizes}
\begin{tabular}{p{1.75in}|p{1.5in}|p{1in}}
\toprule
    Dataset Name    & Kind      & Num. of Tokens \\
\hline
    \textit{PuoData} contents & &  \\
    NCHLT Setswana \cite{eiselen2014developing} 	& Government Documents &  1,010,147 \\
    Nalibali Setswana & Childrens Books & 57,654 \\
    Setswana Bible  & Book(s) & 879,630 \\
    SA Constitution& Official Document  & 56,194 \\
    Leipzig Setswana Corpus BW  & Curated Dataset &  219,149 \\
    Leipzig Setswana Corpus ZA  & Curated Dataset & 218,037 \\
    SABC Dikgang tsa Setswana FB (Facebook)  & News Headlines & 167,119 \\
    SABC MotswedingFM FB & Online Content & 33,092 \\
    Leipzig Setswana Wiki & Online Content  & 230,333 \\
    Setswana Wiki & Online Content  & 183,168 \\
    Vukuzenzele Monolingual TSN  & Government News  & 157,798 \\
    gov-za Cabinet speeches TSN & Government Speeches  & 591,920 \\
    Department Basic Education TSN & Education Material &   708,965 \\
    \hline
    \textbf{PuoData Total} & 25MB on disk   & \textbf{4,513,206} \\
    \hline
\textit{PuoData+JW300} & &  \\
    JW300 Setswana  \cite{agic-vulic-2019-jw300} & Book(s) & 19,782,122 \\
    \hline
    \textbf{PuoData+JW300 Total}   & 124MB on disk  & \textbf{24,295,328}\\
    \hline
\textit{NCHLT RoBERTa Reported\footnote{\url{https://v-sdlr-lnx1.nwu.ac.za/handle/20.500.12185/641?show=full}}} & Mixture & 14,518,437 \\
\hline
\end{tabular}
\end{table}

PuoData is made available as a single dataset for researchers\footnote{\url{https://github.com/dsfsi/PuoData}, \url{https://huggingface.co/datasets/dsfsi/PuoData}}. We now focus on how we built pre-trained models with PuoData and PuoData+JW300.

\subsection{Training the Masked Language Model: {PuoBERTa}}

We first trained two BPE Tokenizers for the PuoBERTa. One tokenizer with PuoData and the other with PuoData+JW300 corpora with 52000 tokens each. The pre-trained masked language models were trained on an NVIDIA Titan RTX with 24 GB of VRAM on an Intel Core i9-9900K with 32GB of RAM. The training information is shown in Table \ref{tab:training-pretraining}.

\begin{table}[ht]
\centering
\caption{Pre-training setup for PuoBERTa and PuoBERTaJW300 (trained with PuoData+JW300) using the Huggingface library.}
\label{tab:training-pretraining}
\begin{tabular}{c | c | c}
\toprule
           & PuoBERTa      & PuoBERTaJW300 \\
\midrule
Epochs          & 100        & 40 \\
Steps           & 525,000   & 1,585,000 \\
Time Training (approx)   & 3 days    & 3 days \\
\bottomrule
\end{tabular}
\end{table}

The pre-trained models are made available for download for research and development\footnote{\url{https://github.com/dsfsi/PuoBERTa}, \url{https://huggingface.co/dsfsi/PuoBERTa}}~\cite{marivate_vukosi_2023_8434796}. We next focus on evaluating the new models on a number of downstream tasks.

\section{Evaluation of PuoBERTa on Downstream Tasks}

The PuoBERTa models were fine-tuned and evaluated on three downstream tasks: Named Entity Recognition (NER), Part of Speech (POS) tagging, and news categorisation/classification. The results showed that the PuoBERTa model achieved state-of-the-art results on all three tasks in terms of F1 score amongst the monolingual models.

\subsection{Evaluation on MasakhaNER}

We first evaluate PuoBERTa on the MasakhaNER 2.0~\cite{adelani2022masakhaner}  benchmark, which is a dataset for named entity recognition in a number of African languages including Setswana. We compare the performance of PuoBERTa to NCHLT RoBERTa and the multilingual models reported in the MasakhaNER 2.0 paper~\cite{adelani2022masakhaner}, namely Afriberta, AfroXLMR-base and AfroXLMR-large models. The results in Table \ref{tab:masakhaner} show that PuoBERTa is very competitive. It does not beat the multilingual models in this case. PuoBERTa+JW300 gets much closer than the rest of the monolingual models.

\begin{table}[ht]
\centering
\caption{Performance of models on the MasakhaNER datasets}
\label{tab:masakhaner}
\begin{tabular}{|l|c|}
\toprule
Model & Test Performance (f1 score) \\
\hline
\textbf{Multilingual Models}~\cite{adelani2022masakhaner}  & \\
AfriBERTa \cite{ogueji-etal-2021-small}     	& 83.2\\
AfroXLMR-base                               				& 87.7\\
AfroXLMR-large                              				& \textbf{89.4}\\
\textbf{Monolingual Models} & \\
NCHLT TSN RoBERTa                        				& 74.2 \\
PuoBERTa                                    					& 78.2 \\
PuoBERTa+JW300                         				& \textit{80.2} \\
\hline
\end{tabular}
\end{table}

\subsection{Evaluation on MasakhaPOS}

For this downstream task we use the same evaluations script as per the MasakhaPOS paper~\cite{dione-etal-2023-masakhapos} to evaluate our PuoBERTa variants plus the NCHLT RoBERTa. The results are shown in Table \ref{tab:masakhapos}. In this test PuoBERTa almost beats the best multilingual model (AfroLM) but PuoBERTa+JW300 beats all the models available.

\begin{table}[ht]
\centering
\caption{Performance of models on the MasakhaPOS datasets}
\label{tab:masakhapos}
\begin{tabular}{|l|c|}
\toprule
Model & Test Performance (f1 score)\\
\midrule
\textbf{Multilingual Models}~\cite{dione-etal-2023-masakhapos}  							& \\
AfroLM \cite{dossou2022afrolm} & \textit{83.8} \\
AfriBERTa	& 82.5 \\
AfroXLMR-base                          					& 82.7\\
AfroXLMR-large                          					& 83.0\\
\textbf{Monolingual Models}             			& \\
NCHLT TSN RoBERTa                      	 		& 82.3 \\
PuoBERTa                               						& 83.4 \\
PuoBERTa+JW300                        				& \textbf{84.1} \\
\bottomrule
\end{tabular}
\end{table}

\subsection{News Categorisation - Daily News}

\subsubsection{Dataset information}

We gather our dataset from the official website of the Government of Botswana's Daily News\footnote{\url{https://dailynews.gov.bw/}}. The website primarily features news content in the English language, while also incorporating some of their materials in Setswana within the designated \textit{Dikgang} segment. Our compiled dataset, denoted as \textit{Daily News - Dikgang}, needed annotation of news categories. Notably, the news items were initially devoid of categorisation information. To rectify this, we embarked on the task of categorising these news items, leveraging the International Press Telecommunications Council (IPTC) News Categories (or codes)\footnote{\url{https://iptc.org/standards/newscodes/}}. For the categorisation process, we employed the highest hierarchical level of the news code taxonomy. Table \ref{tab:iptc_media_codes} provides an overview of these top-level codes, along with their corresponding Setswana translations, which guided our annotation efforts. For this process we involved two expert Setswana annotators who initially annotated the dataset individually and then came together to deal with disagreements on labelling afterwards leading to the final dataset that we use for this paper and also make available. See Appendix A for the Data Statement.  

\begin{table}
\centering 
\caption{International Press Telecommunications Council (IPTC) Media Codes and Setswana Translations}
\label{tab:iptc_media_codes}
\begin{tabular}{| p{0.475\columnwidth} | p{0.475\columnwidth} |}
\hline
         \bfseries English Category 				 		&  \bfseries Setswana Category  \\
\hline
        arts, culture, entertainment and media 		&	Botsweretshi, setso, boitapoloso le bobegakgang  \\
        conflict, war and peace 								&	Kgotlhang, ntwa le kagiso  \\
        crime, law and justice 									&	Bosenyi, molao le bosiamisi  \\
        disaster, accident and emergency incident &	Masetlapelo, kotsi le tiragalo ya maemo a tshoganyetso  \\
        economy, business and finance 					&	Ikonomi, tsa kgwebo le tsa ditšhelete  \\
        education 													&	Thuto  \\
        environment 												&	Tikologo  \\
        health 															&	Boitekanelo  \\
        human interest 											&	Tse di amanang le batho  \\
        labour 															&	Bodiri  \\
        lifestyle and leisure 										&	Mokgwa wa go tshela le boitapoloso  \\
        politics 														&	Dipolotiki  \\
        religion and belief 										&	Bodumedi le tumelo  \\
        science and technology 								&	Saense le thekenoloji  \\
        society 														&	Setšhaba  \\
        sport 															&	Metshameko  \\
        weather 														&	Maemo a bosa  \\
\hline
\end{tabular}
\end{table}

\begin{figure}[ht]
    \centering
    \includegraphics[width=\linewidth]{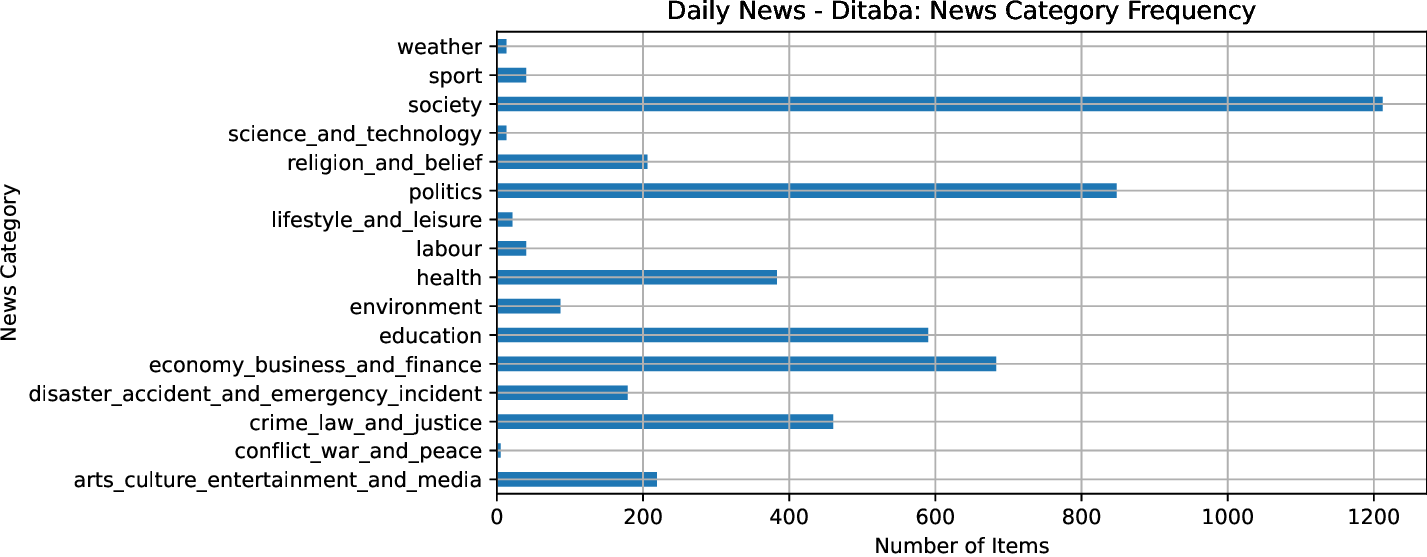} 
    \caption{Frequency of news categories in the \textit{Daily News - Ditaba} dataset}
    \label{fig:ditaba-news-freq}
\end{figure}

Following the annotation process, it is worth noting that not all annotated data was employed in the subsequent evaluation of models for the classification task. Our strategy involved focusing on the top 10 categories, each comprised of a minimum of 80 labelled samples in the full dataset. This selection criterion was informed by the frequency distribution observed within the collected dataset (see Figure \ref{fig:ditaba-news-freq}). As such the final full dataset has a total of 4867 samples,  3893, 487 and 487 samples for train, dev and test splits respectively. 

\begin{figure}
	\begin{center}
		\includegraphics[width=0.6\columnwidth]{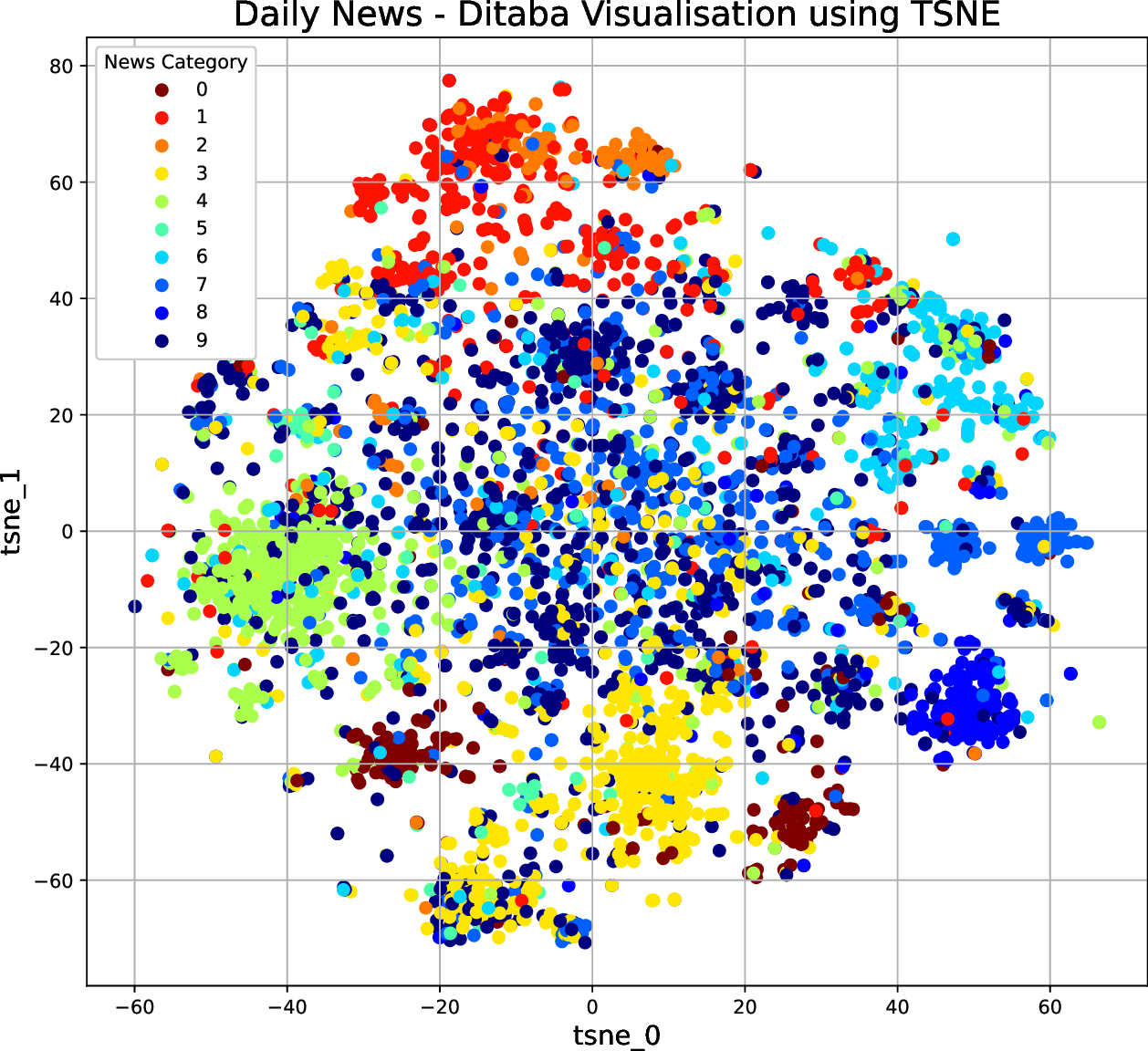}
	\end{center}
\caption{TSNE Visualisation of TFIDF representation of top 10 category news stories. Categories: 0:\emph{arts\_culture\_entertainment\_and\_media}; 
1:\emph{crime\_law\_and\_justice}; 
2:\emph{disaster\_accident\_and\_emergency\_incident};
3:\emph{economy\_business\_and\_finance};
4:\emph{education};
5:\emph{environment};
6:\emph{health};
7:\emph{politics};
8:\emph{religion\_and\_belief};
9:\emph{society}.}
\label{fig:tfidf_tsne_dailynews}
\end{figure}

In order to delve into the textual data, we present a 2-dimensional visualisation of the news dataset in Figure \ref{fig:tfidf_tsne_dailynews}, where each category is represented by a unique color. This visualisation was constructed by applying a T-distributed Stochastic Neighbour Embedding (TSNE) \cite{van2008visualizing} model to the TF-IDF representation of the news corpus data. The visualisation gives us an initial view of the distribution of the data and how separable the categories might be.  We make the complete dataset available alongside this paper\footnote{\url{https://github.com/dsfsi/PuoBERTa}} \cite{marivate_vukosi_2023_8434796}, fostering openness and reproducibility.
 
\subsubsection{Training and performance}

For the news categorisation task, we performed 5-fold cross-validation using the different models at hand. We performed the cross validation by combining the train and dev sets and then doing an 80/20 (train/validate) split 5 times randomly. We compared PuoBERTa, PuoBERTaJW300 to baselines of the NCHLT RoBERTa Model, and also logistic regression with TFIDF.  We use the TFIDF+logistic regression to provide an insight on how well the more powerful models perform as well as make sure we have a more efficient baseline. We further provide the performance on the test set for the models we have. For all the RoBERTa-based models, the fine-tuning was at 5 epochs. The results of our classification task experiments are shown in Table \ref{tab:daily_news_results}.
\begin{table}[ht]
\centering 
\caption{Setswana News Categorisation Task Results}
\label{tab:daily_news_results}
\begin{tabular}{| l | c | c|}
\hline
       \bfseries Model & \bfseries 5-fold Cross Validation F1 &  \bfseries Test F1  \\
\hline
      Logistic Regression + TFIDF  		 & 60.1			         & 56.2	         \\
      NCHLT TSN RoBERTa  		             & 64.7	                 & 60.3  \\
      PuoBERTa 							 & 63.8 	             & 62.9   \\
      PuoBERTaJW300 					     & \textbf{66.2} 	     & \textbf{65.4}   \\
\hline
\end{tabular}
\end{table}

\begin{figure}
	\begin{center}
		\includegraphics[width=0.75\columnwidth]{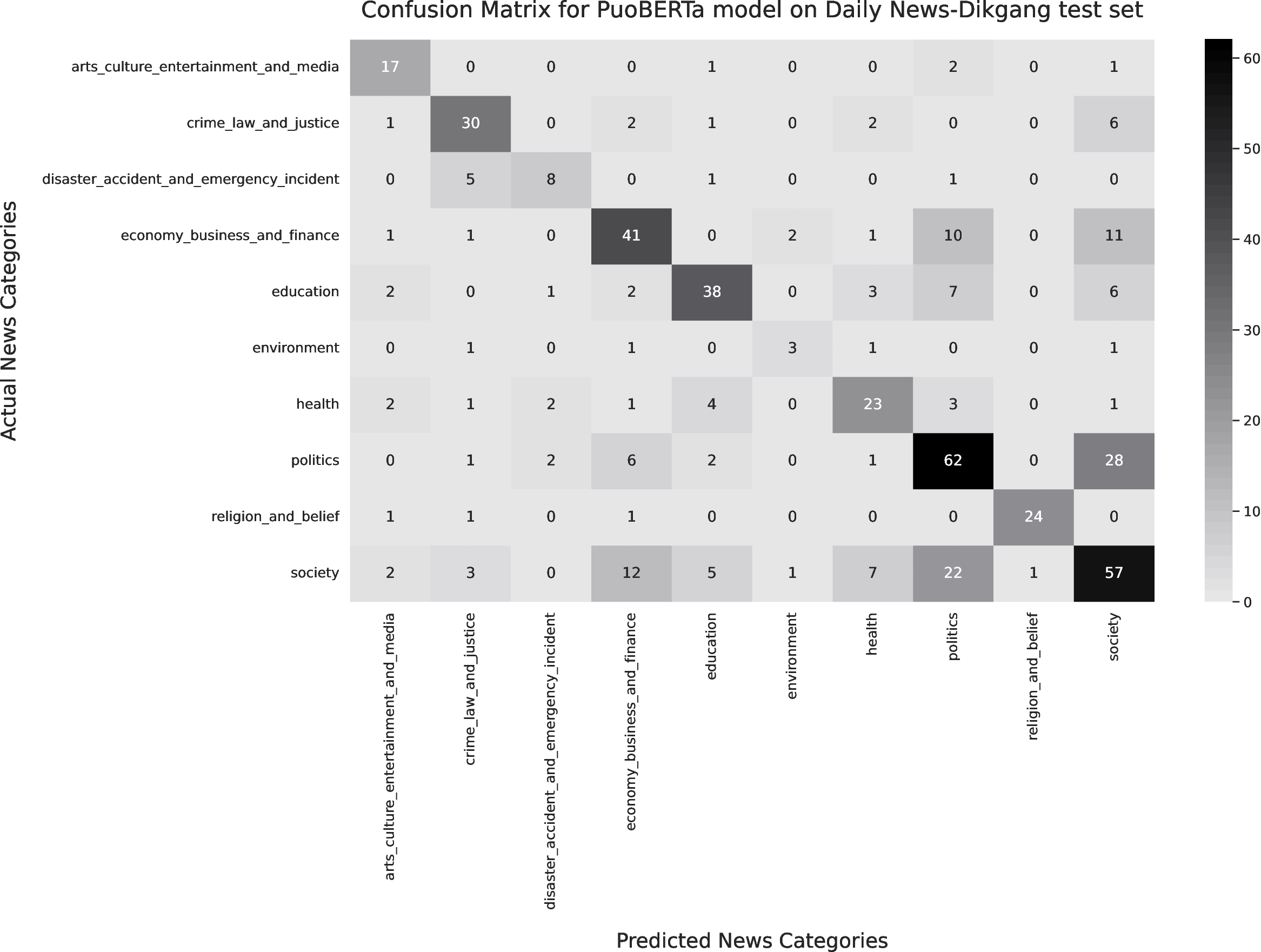}
	\end{center}
\caption{Confusion Matrix PuoBERTa Model}
\label{fig:daily_news_ditaba_confusion}
\end{figure}

There is a performance improvement by using the RoBERTa based models. The performance improvements by PuoBERTa and PuoBERTaJW300 are even better respectively. To note, the logistic regression TFIDF vectorizer is only trained using the news training data while the RoBERTa tokenizers have not been specifically trained using the Daily News data. It is still an interesting result as development of the news dataset is also useful for less powerful models.  To better understand the classification performance, we took the best performing model of the PuoBERTaJW300 and show its confusion matrix in Figure \ref{fig:daily_news_ditaba_confusion}. We observe in this confusion matrix that the models makes more mistakes when trying to predict \textit{economy\_business\_finance}, \textit{politics}, \textit{society} categories. 

\subsection{Observations Given the Downstream Tasks}

PuoBERTa in general is a competitive model. When we include JW300 data (PuoBERTaJW300), the model improves further and performs as state of the art for the downstream tasks. This brings up a question of what is lost by not being able to use JW300 as a language resource for Setswana and other low resource languages covered by JW300? The performance of PuoBERTa is close to that of PuoBERTaJW300 and is reassuring as it would allow more researchers to focus on building better African language corpora.  

\section{Conclusion and Future Work}

In this paper, we presented the creation of a number of artifacts as we pursued creating a state of the art pre-trained Language Model for the Setswana Language. We were able to gather a corpora (of diverse sources) named PuoData that we release along with this paper with a permissive license. We further discuss the PuoData+JW300 corpora which allows us to measure the impact of pretraining with JW300 included in the corpora. For the POS, NER and a new news categorisation task (\emph{Daily News - Dikgang}) we show that PuoBERTa and PuoBERTaJW300 are competitive language models that have state of the art or near state of the art performance in those downstream tasks. 

We look forward to seeing how the community uses this model and different application areas. There are a few directions to still be taken that we could not cover in this paper. The corpora needs to be further analysed for bias \cite{2023twi,2022bias,haddow-etal-2022-survey} and the models evaluated for challenges with domain shift \cite{adelani-etal-2022-thousand}. We have shown that if we include JW300, the corpora size increases 4 fold, but at what cost? Further in comparison with large multilingual models, PuoBERTa and PuoBERTaJW300 are 300 MB (200MB compressed) might benefit from further optimisations that may make it more useful for low resource scenarios (compute, access to internet etc.). 

For the news categorisation task, there could be more work done to understand how both the simple logistic regression model and other more complex ones can benefit from more feature or label engineering of the dataset to combine labels that might be two similar in hindsight. We also suggest looking into how we can standardise datasets such as \cite{ifeoluwa2023masakhanews} etc. to all use a standard categorisation set to enable better transfer learning and evaluation for news. 

\section{Acknowledgement}

We want to acknowledge the feedback received from colleagues at DSFSI at Univeristy of Preotria and Lelapa AI that improved this paper. We would like to acknowledge funding from the ABSA Chair of Data Science, Google and the NVIDIA Corporation hardware Grant. We are thankful to the anonymous reviewers for their feedback.

\bibliographystyle{splncs04}
\bibliography{references_puoberta}

\newpage
\section*{Appendix: Data Statement for Daily News - Dikgang Categorised News Corpus}

\noindent\rule{12.1cm}{0.4pt}\\
\emph{Dataset name:} Daily News - Dikgang Categorised News Corpus\\
\emph{Citations:} Cite this paper.\\
\emph{Dataset developer(s):} Vukosi Marivate (vukosi.marivate@cs.up.ac.za) and Valencia Wagner (valencia.wagner@spu.ac.za)\\
\emph{Data statement author(s):}  Vukosi Marivate\\
\emph{Organisation:} Data Science for Social Impact Research Group \\ \url{https://dsfsi.github.io}),\\ Department of Computer Science, University of Pretoria, South Africa\\ and Sol Plaatje University \\
\noindent\rule{12.1cm}{0.4pt}

\subsection*{A. CURATION RATIONALE}

The motivation for building this dataset was to provide one of the few annotated news categorisation datasets for Setswana. The task required identifying a high-quality Setswana news dataset, collecting the data, and then annotating leveraging the International Press Telecommunications Council (IPTC) News Categories (or codes)\footnote{\url{https://iptc.org/standards/newscodes/}}. The identified source was the Daily News\footnote{\url{https://dailynews.gov.bw/}} (Dikgang Section) from the Botswana Government. All copyright for the news content belongs to Daily News. We collected 5000 Setswana news articles. The distribution of final categories for the dataset are shown in Figure \ref{fig:ditaba-news-freq}.

\subsection*{B. LANGUAGE VARIETY}

The language of this data set is Setswana (primarily from Botswana). 

\subsection*{C. SPEAKER DEMOGRAPHIC}

Setswana is a Bantu languages that is spoken in Botswana as well as several regions of South Africa \cite{palai2004word}.

\subsection*{D. ANNOTATOR DEMOGRAPHIC}

Two annotators were used to label the news articles based on the \emph{Daily News - Dikgang} news. Their deomographic information is shown in Table \ref{tab:annot}.

\begin{table}[ht]
	\caption{Annotator demographic}
	\centering{
	\begin{tabular}[h]{p{1in}|p{1in}|p{1in}}
		&\bfseries 1 & \bfseries 2 \\
		\hline
		Description&Annotator&Annotator\\
		\hline
		Age&30-35&20-25\\
		\hline
		Gender&Female&Male\\
		\hline
		Race/ethnicity&Black/African&Black/African\\
		\hline
		First Language(s)&Setswana&Setswana and isiXhosa\\
		\hline
		Linguistics training & Often works as a Setswana - English interpreter& Studied linguistic anthropology and works as a translator/interpreter\\
		\hline
	\end{tabular}
        }
	\label{tab:annot}
\end{table}

\subsection*{E. PROVENANCE APPENDIX}

The original data is from the Daily News news service from the Botswana Government. 

\end{document}